\documentclass[10pt,twocolumn,letterpaper]{article}

\usepackage[pagenumbers]{wacv} 

\usepackage{graphicx}
\usepackage{amsmath}
\usepackage{amssymb}
\usepackage{booktabs}

\usepackage[pagebackref,breaklinks,colorlinks]{hyperref}

\usepackage[capitalize]{cleveref}
\crefname{section}{Sec.}{Secs.}
\Crefname{section}{Section}{Sections}
\Crefname{table}{Table}{Tables}
\crefname{table}{Tab.}{Tabs.}

\begin{document}

\title{Deep Model Interpretation with Limited Data : A Coreset-based Approach}

\author{Hamed Behzadi-Khormouji \hspace{1cm} José Oramas\\
University of Antwerp, sqIRL/IDLab, imec\\
Antwerp, Belgium}

\maketitle

\begin{abstract}
   Model Interpretation  aims at the extraction of  insights from the internals of a trained model. A common approach to address this task is the characterization of relevant features internally encoded in the model that are critical for its proper operation. Despite recent progress of these methods, they come with the weakness of being computationally expensive due to the dense evaluation of datasets that they require. As a consequence, research on the design of these methods have focused on smaller data subsets which may led to reduced insights. To address these computational costs, we propose a coreset-based interpretation framework that  utilizes coreset selection methods to sample a representative  subset of the large dataset for the interpretation task. Towards this goal, we propose a similarity-based evaluation protocol to assess the robustness of model interpretation methods towards the amount data they take as input. Experiments considering several interpretation methods, DNN models, and coreset selection methods show the effectiveness of the proposed framework.
\end{abstract}


\section{Introduction}
\label{sec:intro}
Over the last few years, the quest to understand the inner workings of deep models, particularly Convolutional Neural Networks (CNNs), has gained significant attention~\cite{tcav, vebi,ace,topicmodel,ice}. This task, referred to as \textit{model interpretation}, have frequently  been addressed by the extraction of  insights from relevant features  internally encoded by the model~\cite{vebi}.

The interpretation capabilities can be added to the models either during the training phase, leading to models that are  interpretable-by-design~\cite{protopnet,protopshare,prtotree}, or on top of a pretrained model, i.e. in a post-hoc manner~\cite{ace,vebi,topicmodel,ice}. These post-hoc methods have, on the one hand,  the advantage of not interfering with the training phase of the original model. 
On the other hand, despite the effectiveness of these interpretation methods at providing insights on the internal encoded representations, they face a significant challenge. They suffer from a high computational cost due to the requirement of analysing the complete training dataset or third-party datasets specially designed for the task~\cite{networkdissection}. This becomes more critical in the case of large datasets like ImageNet~\cite{imagenet}. Consequently, the majority of these methods have focused their design and evaluation on small datasets~\cite{linear_probing,tcav,ace,selectivity_index,topicmodel,acts_mining2023}. Given the growing complexity and size of deep models, interpreting them using the entire dataset becomes increasingly impractical and computationally expensive, which usually makes it unaffordable by startups and non-profit organizations.

\begin{figure}[t]
  \centering
   \includegraphics[width=0.98\linewidth]{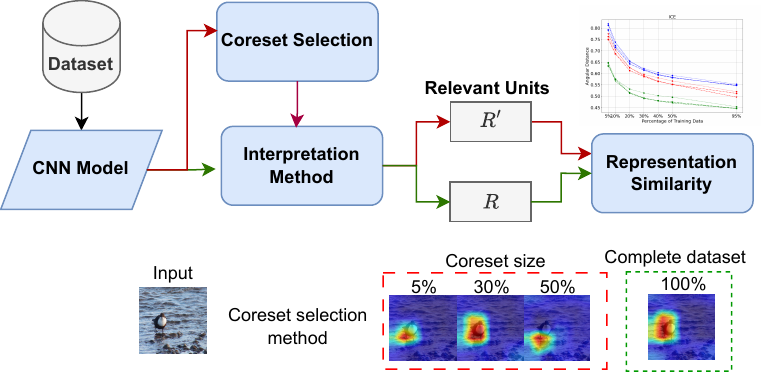}
   \caption{Proposed is a coreset-based model interpretation framework. The CNN model computes activation maps for the entire dataset, then a coreset selection method picks representative subsets. Using these subsets and entire activation maps, an interpretation method obtains relevant internal features encoded by the CNN model. Finally, two evaluations compare the quality of obtained features from the coreset.}
   \label{fig:teaser}
\end{figure}

From the training point of view, such massive deep models containing millions of training parameters require a high computational resources. To cope with this problem, the  \textit{Coreset selection} research line has recently emerged as a promising technique to reduce computational costs~\cite{deepcore}. \textit{Coreset selection} aims at selecting a small informative subset of samples from a large dataset, that when used to train a model, leads to performance comparable to that obtained with the complete dataset~\cite{d2pruning}. Following this goal, the design of \textit{Coreset selection} methods has prioritized the maximization, or at least preservation, of classification accuracy~\cite{deepcore,d2pruning,moderate_selection,paul2021deep}. 

\textbf{Our Proposal.} 
To address the computational challenges of model interpretation methods, we propose a model interpretation framework that conducts a coreset selection pre-processing step prior to the interpretation task. We complement this  with a similarity-based evaluation protocol aimed at assessing the robustness of model interpretation methods to the amount of data they are fed as input (Fig.~\ref{fig:teaser}).

More specifically, in the first part of the framework, we begin by applying coreset selection methods to the entire dataset to select representative samples. %
Next, using this subset, we apply an interpretation method to a pretrained model to identify/learn relevant features critical for the functioning of the model. 
To the best of our knowledge, this is the first time coreset selection is integrated in the context of speeding-up the model interpretation task.

In the second part of the framework, we aim to assess the quality of the relevant features obtained from the data subset produced by the coreset selection method. Towards this, we measure the similarity between the representation of the relevant features obtained using the complete dataset and that of the relevant features extracted by using the coreset. This is different from the coreset selection research line, which commonly focuses solely on classification metrics.
The rationale behind measuring representation similarity is to obtain a coreset that closely reflects the interpretation insights obtained from the full  dateset.  Moreover, this leads to investigate stability of interpretation method in providing insights using different data budget, that are similar to those insights obtained from complete dataset. This is also complementary to existing interpretation evaluation protocols~\cite{networkdissection,net2vec,unified_protocol}, where an external annotation dataset is needed to measure coverage of semantic concepts.


\section{Related Work}
\label{sec:related_work}
\noindent We position our work w.r.t. the following three aspects:

\textbf{Interpretation Methods.} In this paper, we focus on post-hoc interpretation methods, wherein the interpretation method is applied after the training phase. While post-hoc methods do not interfere with the training phase, they require additional time and computational resources to analyze the model. 
For example, Visual Explanation By Interpretation (VEBI) \cite{vebi} considers the average of activations computed by every single unit, i.e. convolutional filter, from every image in the training set. 
~\cite{ace} applies a clustering approach to the activations of a given convolutional layer, computed from the hundred patches extracted from  each image of a class at a time. This inherently results in a high computational costs and time consumption. 
Invertible Concept-based Explanations (ICE)~\cite{ice} operates at the activations of a given class, one class at a time. Topic-based interpretation~\cite{topicmodel} integrates an interpretable module defining activation topics from the model. Learning topics requires passing images through the model for several epochs. 
\cite{subnetworks} proposes the use of a binary classifier to identify internal units, i.e. convolutional filters, specific for the given class. This requires to repeat the method for each class involving the complete dataset every time. Consequently, reproducing the results from this method is highly expensive. 
With the exception of \cite{vebi,ice,concept_attribution, subnetworks,revers_probing} that have considered the complete ImageNet dataset, a good amount of the existing post-hoc interpretation methods, due to their computational costs, have focused on either small datasets~\cite{topicmodel,pace, linear_probing, intervention,acts_mining2023} or a small subset of ImageNet~\cite{ace, tcav,selectivity_index}. Here we aim to adopt coreset selection as a pre-processing step to alleviate the computational costs inherent to interpretation methods. At the same time, we focus on preserving the capability of extracting insights with comparable quality to those extracted when considering large datasets.


\textbf{Coreset selection.} There are two common approaches to choose samples from a dataset as a coreset~\cite{deepcore}. (1) Model uncertainty-based methods that aim to assign a score to the samples based on the training dynamics. (2) Geometry-based methods that aim to maximize data diversity in the coreset. 
The former group, uncertainty-based methods, considers different criteria related to training phase as a score function. Considered criteria include the number of times a given sample is misclassified during the training phase~\cite{toneva2018empirical}, entropy of model prediction vectors~\cite{coleman2020selection} or the $l2$-norm of model error vectors~\cite{paul2021deep}.
In contrast, geometry-based methods operate on the assumption that samples closely located in the representation space offer redundant information. Consequently, the selection criteria in this category rely on factors such as minimizing the largest distance from any data point to its k-nearest center~\cite{sener2018active}, largest distance from centers in k-means clustering of sample activations~\cite{sorscher2022beyond} or median distance from the cluster center in Moderate Selection~\cite{moderate_selection}. Recently, D2Pruning~\cite{d2pruning} proposed to combine both aspects, i.e. uncertainty-based and geometry-based selection methods, to construct coresets.
Here, we consider the geometry-based approach for our interpretation pipeline for two reasons. First, the uncertainty-based approach relies on the training phase. That is to say, they need to train the model from scratch to collect/consider training dynamics, while we focus on post-training (post-hoc) interpretation. 
Second, our integration of coreset selection targets the extraction, at a low computational cost, of insights similar to those obtained when using the complete dataset. Thus, maximizing the extraction of representative samples. This differs from the original goal of coreset selection, which aims to select samples maximizing model performance, e.g. accuracy. It is for this reason that we take into account the representation similarity between the interpretations provided by coreset and the full dataset.  
%


\textbf{Evaluation of interpretation methods.}
Some works have evaluated the feedback produced by their interpretation methods through user studies. Due to the subjective nature of these studies, reproducing their results is somewhat  problematic~\cite{ace,pace}. 
In the case of quantitative evaluation of model interpretation methods, some methods track model output dynamics either accuracy~\cite{linear_probing,vebi,revers_probing,conceptwhitening,tcav} or errors~\cite{network_inversion} without considering the similarity of internal representations. 
The other group measures the semantic alignment between the produced interpretation feedback and annotation masks from semantic concepts~\cite{networkdissection,net2vec,unified_protocol}. These evaluation protocols require an additional dataset; usually required to be annotated at the pixel level. Moreover, they require the dense application of this dataset for the measurement of coverage and the estimation of specific parameters, e.g. activation threshold~\cite{networkdissection}. 

Regarding representation similarity, there are extensive works that measure similarity between internal representations from different models~\cite{wang2018towards,raghu2021vision,nguyen2020wide,davari2023reliability}. However, these are beyond the scope of model interpretation as they investigate general means to measure similarity across representations.
More related to the interpretation task, \cite{tcav} utilizes cosine similarity to sort images using learned Concept Activation Vectors (CAV), i.e. learned relevant features. To do this, it computes the activation maps from a given convolutional layer for each image. Next, the cosine similarity between activation maps and the learned CAVs is computed. Finally, the images are sorted based on their similarity score. In this procedure, \cite{tcav} assumes each activation map to be semantically aligned with each element of CAV. However, in our case, different data budgets are used to extract relevant features (the source of the insights) via the interpretation method. Hence, there is no guarantee that the relevant features obtained through the coreset are  aligned with those from entire dataset. 
Taking this into account, different from the discussed methods, we consider an evaluation protocol that considers a representation similarity of relevant features produced by an interpretation method. As part of this, we propose an evaluation protocol for model interpretation methods that measure the stability of their output w.r.t the size of the dataset that is fed to it.
%
%


\section{Coreset-based model interpretation}

This section describes the procedure of model interpretation via coreset selection method in three parts: (1) Description of the proposed interpretation pipeline, (2) modifications done on the considered coreset selection algorithm, and the (3) proposed similarity-based evaluation protocol.

\subsection{Proposed interpretation pipeline}
Consider $\boldsymbol{D}{=}\{\boldsymbol{X}^i,\boldsymbol{Y}^i\}_{i=1}^n$ a dataset containing $n$ image samples $\boldsymbol{X}^i$ and their corresponding class label $\boldsymbol{Y}^i$. Also, consider $\boldsymbol{C}$ as number of classes, and $\boldsymbol{A}^i \in \mathbb{R}^{w\times h\times d}$ as the activation maps produced from a given convolutional layer, in our case the last convolutional layer, from pushing sample $\boldsymbol{X}^i$ into the  model. $h$, $w$, and $d$ indicate the height, width, and depth of the activation map, respectively. %
We apply an average global pooling operation on each channel of the activation maps $\boldsymbol{A}^i$ to form a new activation vector $\boldsymbol{A'}^i \in \mathbb{R}^{1\times d}$. 
Next, a coreset selection method $\boldsymbol{CS}$ with a selection rate $\rho$ is applied on the activation vector of each separate class to form a coreset dataset $\boldsymbol{D'} = \{\boldsymbol{X}^j,\boldsymbol{Y}^j\}_{j=1}^m$ where $m < n$. 
Then, an interpretation method $\boldsymbol{I}$ is applied on the model by only considering the samples in $\boldsymbol{D'}$ which results in insights extracted from a set of detected relevant internal features $\boldsymbol{R'} \in \mathbb{R}^{p\times q}$. $p$ indicates the number of extracted relevant features, each having a dimensionality $q$. 


\subsection{Coreset selection method modification}
While our interpretation framework allows for any geometry-based coreset selection, we have considered two recent proposed coreset selection methods namely Moderate Selection~\cite{moderate_selection} and our modification of D2Pruning~\cite{d2pruning}.

\textbf{DGPruning}. D2Pruning combines aspects of both geometry-based and un-certainty-based approaches. 
It constructs a graph of the dataset, where each node represents a data sample, and edges between nodes are determined by the Euclidean distance between their corresponding activation maps computed from the model. The initial score of each node is determined based on the collected training dynamics namely \textit{Area Under the Margin} (AUM) which measures the probability gap between the target class and next largest class across all training epochs. 
The selection procedure is determined by updating the nodes' score by forward and backward message transfer among nodes as formulated in~\cite{d2pruning}.
However, since our interpretation framework takes a pretrained model as input, we adapt this selection method to operate without the need  of training dynamics. Therefore, drawing inspiration from \cite{moderate_selection}, we determine the initial node scores based on the Euclidean distance of their activation maps to the class center. The class center is computed by averaging the activation maps belonging to the given class. The rest of the selection procedure remains unchanged. We refer to this modification as Diversity-Graph based Pruning (DGPruning). 

\subsection{Coreset-based evaluation protocol}
\label{sec:eval_protocol}

For our method to be effective, the insights extracted by the interpretation method when  considering the coreset should be similar to those obtained from the complete dataset. In practice, this is achieved by measuring the similarity between the representations encoded by the internal relevant features $\boldsymbol{R}$ and $\boldsymbol{R'}$ extracted from both the full dataset and the coreset.

Towards this goal, we compute activation maps $\boldsymbol{A}^{c} {\in} \mathbb{R}^{n^c \times w \times h \times d}$ from last convolutional layer, pertaining to all $n^c$ samples from class $c$.
Given $\boldsymbol{A}^{c}$, we employ both $\boldsymbol{R}$ and $\boldsymbol{R'}$ to compute interpretation maps $\boldsymbol{Z}^{c}$ and $\boldsymbol{Z}'^{c}$ with the shape $n^c {\times} w' {\times} h' {\times} d'$. The computation of these interpretation maps is done according to the protocol of each interpretation method (Sec.~\ref{sec:framework_setup}). We reshape the interpretation maps into two dimensions $(n^c.w'.h'){\times} d'$. 
Since the number of samples and data distributions between the coreset and the full dataset are different, there is no guarantee that interpretation elements in both $\boldsymbol{R}$ and $\boldsymbol{R'}$ are aligned; the same holds between $\boldsymbol{Z}^{c}$ and $\boldsymbol{Z}'^{c}$. Hence, we consider an alignment-based similarity metric that finds an optimal transformation to align two representations. To do so, Partial-Whitening Shape Metric~\cite{williams2021generalized}, denoted as $\phi(\bullet)$, is applied for measuring the angular distance between $\boldsymbol{Z}^{^c}$ and $\boldsymbol{Z'}^{^c}$. This procedure is done for $C$ classes and similarity is reported as the average across classes (Eq.\ref{eq:average_angular_dist}). The angular distance is bounded in the interval $[0, \pi]$~\cite{williams2021generalized}. As a result, a lower angular distance indicates highest similarity. 


\begin{equation}
\label{eq:average_angular_dist}
    \boldsymbol{\Phi} = \frac{1}{C} \sum_{c=1}^C \phi(\boldsymbol{Z}^{^c},\boldsymbol{Z'}^{^c})
\end{equation}


\textbf{Robustness to dataset size.} 
The methodology described above, touches on an aspect of model interpretation that have received close to no attention in the literature: their sensitivity towards the amount of data they require as part of their operation. In an ideal scenario,  the size of the dataset used for model interpretation should not have a critical effect on the extracted insights given that the distribution of the representation space of the dataset is preserved. 

To enable the measurement of the sensitivity of a given interpretation method w.r.t. this aspect, we  propose an extension of the procedure described above. More specifically, we apply Eq.~\ref{eq:average_angular_dist} in a systematic manner for coresets of varying size. Thus, focusing on the difference between representations from the insights $\boldsymbol{Z}^c$, obtained from the complete dataset and $\boldsymbol{Z}'^c_{\rho}$, obtained from a coreset of a proportional size $\rho {\in} \{10\%, 20\%,\dots, 50\%\}$.
Finally, the robustness of a given interpretation method can be measured as the mean and standard-deviation of $\boldsymbol{\Phi}$ (Eq.~\ref{eq:average_angular_dist}) across multiple coreset sizes. 


\section{Evaluation}
\subsection{Interpretation framework setup}
\label{sec:framework_setup}
\textbf{Interpretation methods, CNNs, and dataset}. We consider three interpretation methods to evaluate the proposed framework.  Namely VEBI~\cite{vebi}, ICE~\cite{ice}, and Topic-based interpretation~\cite{topicmodel}; the hyperparameters used for these methods are detailed in the Supplementary Material. These methods have been applied to VGG19~\cite{vgg19}, Resnet18, and Resnet50~\cite{resnet50}, pretrained on ImageNet~\cite{imagenet}, and are available in Pytorch~\cite{pytorch}. Model interpretation methods, i.e. the identification/learning of relevant features, have been conducted on the training set from ImageNet, and the evaluation has been performed in the validation set.

\noindent\textbf{Coreset selection methods}. We consider two geometry-based selection methods: (1) Moderate Selection~\cite{moderate_selection} and (2) DGPruning, our extension of D2Pruning \cite{d2pruning}. Moderate Selection measures the average of activations pertaining to samples of a given class as the representative center of that class. 
This is followed by the measurement of the Euclidean distance between each activation map from class $c$ and the class center. These distances are sorted in a list. 
Next, samples whose distances are close to the median of the distances are considered as part of the coreset. 

Different data selection ratios $\rho$ are considered for the experiments such as $5\%$, $10\%$, $20\%$, $30\%$, $40\%$, and $50\%$. Besides, we have defined two different baselines. First, \textit{Random} selection, wherein samples are selected randomly, as a baseline for geometry-based selection methods. Second, $95\%$ of the dataset as a baseline to investigate the quality of obtained relevant features by the interpretation methods on the lower percentage selections.

\noindent\textbf{Interpretation maps computation}. We employ the obtained relevant features, i.e., $\boldsymbol{R}$ and $\boldsymbol{R'}$ by each interpretation method on the activation maps to compute interpretation maps $\boldsymbol{Z^c}$ and $\boldsymbol{Z'^c}$, respectively. This operation is performed according to the protocol of each interpretation method. Since ICE~\cite{ice} and Topic-based interpretation~\cite{topicmodel} were originally applied to the last convolutional layer, the activation maps from this layer are modified by applying the learned relevant features. 
In the case of ICE, it reduces the depth of the activation maps using learned components (learned relevant features), followed by inverting the result into the activation maps dimensionality, considering as interpretation maps. 
In the case of Topic-based interpretation, activation maps are multiplied by the learned topics (learned relevant features). The resulted maps are considered as interpretation maps. 
Different from these, VEBI considers the activation maps from all layers and identifies indices of the convolutional filters, i.e. internal units, that are indicators of the classes of interest.
Then, the concatenation of the activation maps per layer produced by the identified relevant units from the same layer are considered as the interpretation maps.


\subsection{Assessing the relevance of the extracted  features}
\label{sec:acc_intrpr}
This experiment aims to assess whether the features obtained by the interpretation methods via coresets are indeed relevant to the model being analyzed. To achieve this, we measure model classification accuracy by paying close attention to the  identified/learned relevant features.

Given a validation set, each image of the validation set is passed through the model. Activation maps are modified by considering the relevant features in accordance with the descriptions in Sec.~\ref{sec:framework_setup}. 
Then, these modified activations, considered as interpretation maps, are passed through the classifier part of the CNN model to make a prediction. For the case of VEBI, activation maps from the identified relevant units are perturbed by setting them to zero. Then, the possible drop in the classification accuracy is tracked to verify the relevance of the identified units to their corresponding class. The assumption is that relevant features when suppressed would lead to a significant drop in accuracy.

\begin{figure*}[t]
  \centering
   \includegraphics[width=\linewidth]{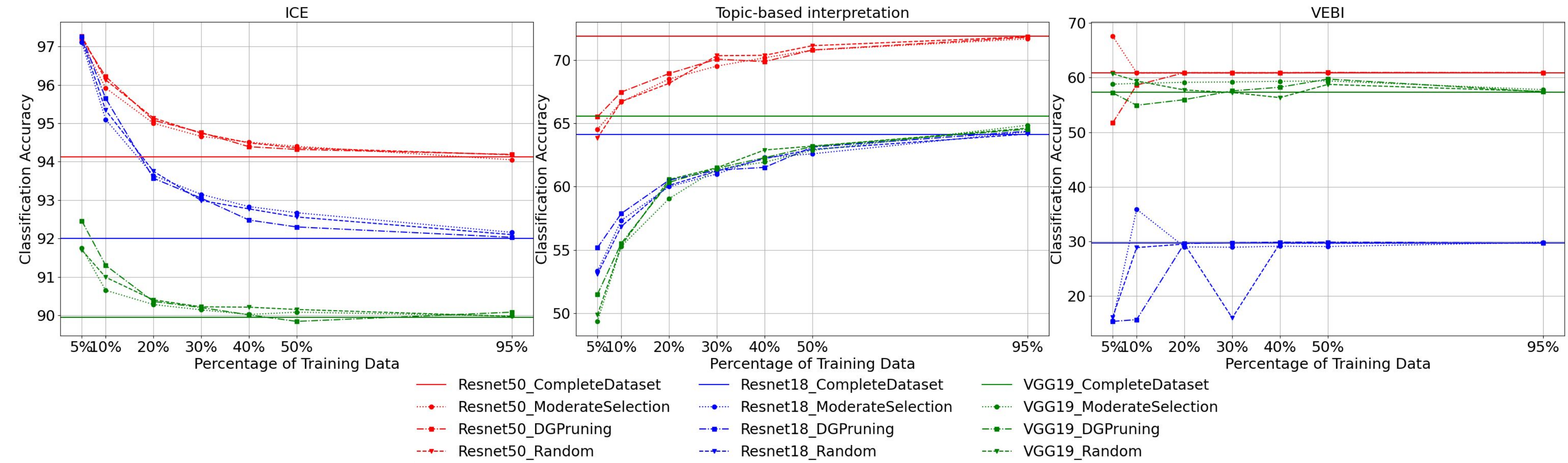}
   \caption{Classification accuracy based on relevant features by different combinations of interpretation and coreset selection methods.}
   \label{fig:classification_accuracy}
\end{figure*}

Fig.~\ref{fig:classification_accuracy} presents the classification accuracy based on the obtained relevant features by ICE (left), Topic-based interpretation (middle), and VEBI (right). In each plot, the color represents the considered CNN models and the marker the different coreset selection methods. For reference, we have also reported classification accuracy (solid lines) when considering the complete dataset (coreset size at $100\%$) in the interpretation procedure.
For all three interpretation methods, the closer the accuracy curve from the coreset-based method is to that from the complete dataset, the better. 

Several observations can be made from these results. Firstly, as could be expected, increasing the data budget in the coreset selection leads to a classification accuracy that is closer to that of complete dataset. 
Secondly, the classification accuracy from lower percentages of data is comparatively close to that of full data, suggesting that coreset selection methods provide effective results when using large datasets. 
Thirdly, among the applied coreset selection methods, random selection yields very competitive results compared to geometry-based coreset selection. This suggests that random selection serves as a strong baseline, specially considering its reduced computational costs, and could be further explored for improving coreset selection methods. 
Fourthly, results from ICE show that decreasing the data budget in the coreset selection leads to a higher accuracy. We believe 
this might be related to its dependence on NMF~\cite{fevotte2011algorithms}. In this regard,~\cite{nmf_survey} has stressed that while NMF is accurate for small datasets, its performance drops when scaling to large datasets.
Finally, considering a very small data budget, for example, $5\%$, the results from ICE are closer to those from full data compared to those from VEBI and Topic-based interpretation. Given $5\%$ data budget, the average of accuracy distance to that of complete dataset for ICE, VEBI, and Topic-based interpretation is $3.4\%$, $7\%$, and $10.93\%$, respectively. 

In summary, the obtained results indicate that even when the amount of data fed as input to the interpretation method is reduced (via coreset selection), the extracted features are still very relevant to the classification task.

\begin{figure}[t]
  \centering
   \includegraphics[width=\linewidth]{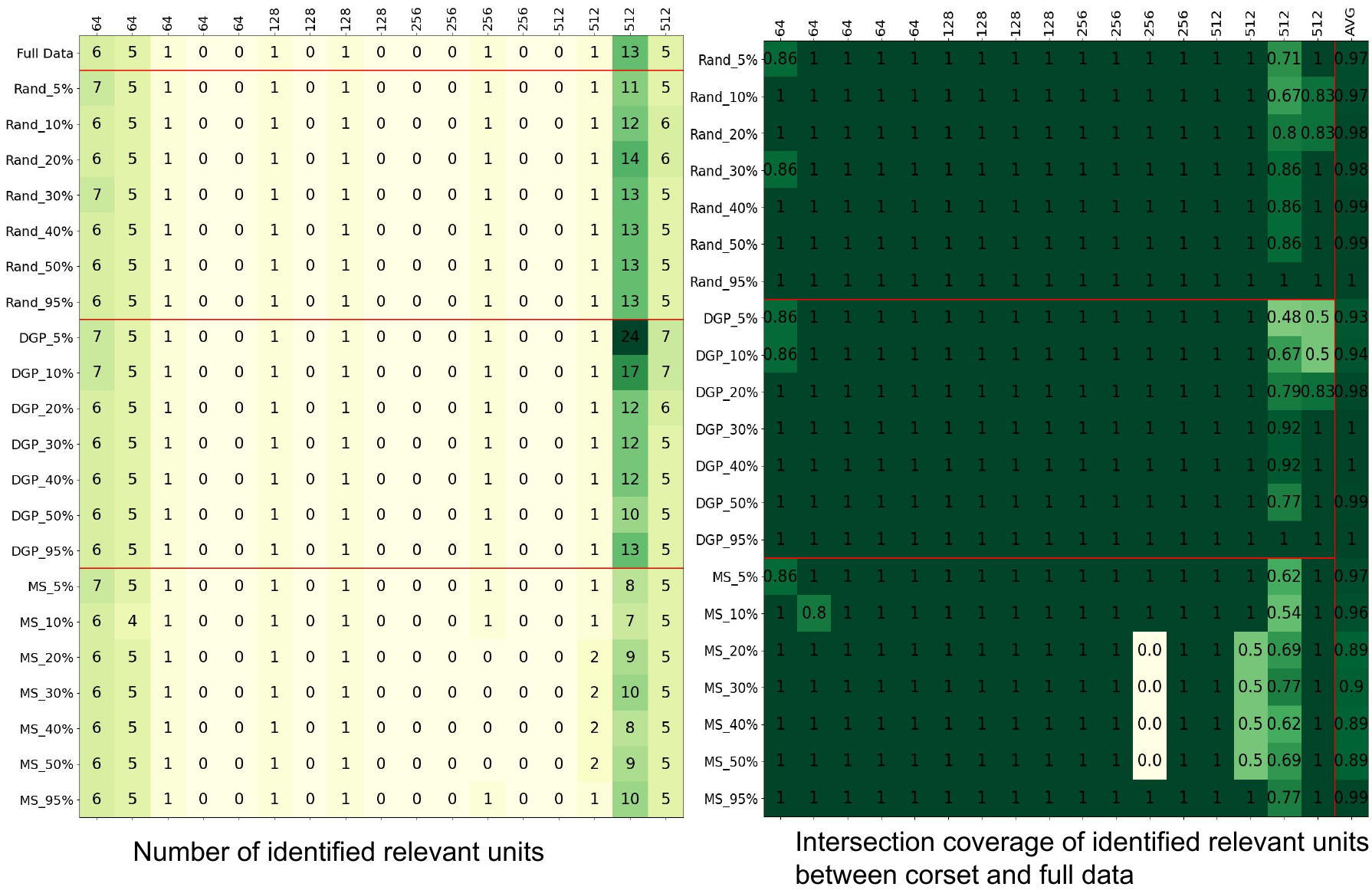}
   \caption{An analysis of identified units by VEBI using different coreset selection methods on Resnet18. \textbf{Left} shows number of identified units per layer. \textbf{Right} shows coverage of similar identified units between each coreset selection method and complete dataset. Rand, DGP, and MS stand for Random, DGPruning, and Moderate Selection methods.}
   \label{fig:vebi_unitindices_analysis}
\end{figure}


\subsection{Coreset-based relevant features similarity} 
\label{sec:intrfeedbacksim}
This section investigates whether the relevant features output of the interpretation method when considering the coresets have similar characteristics to those obtained from complete dataset. 
To do so, we report the  similarity of the representations encoded by the relevant features obtained from the coreset and full data following the protocol described in Sec.~\ref{sec:eval_protocol}. 
Since ICE and Topic-based interpretation focus on the last convolutional layer, we report the measured similarity from the relevant features from this layer. 
Worth noting is that while for these two methods the  interpretation maps have the same depth $d'$, independently of the data source ($\boldsymbol{Z^c}$ from the complete dataset or $\boldsymbol{Z'^c}$ from the coreset), VEBI may not identify the same number of relevant units under different data budgets. As a consequence, the depth of interpretation maps ($\boldsymbol{Z^c}$) and ($\boldsymbol{Z'^c}$) might be different when considering VEBI.

To address this disparity, we subsample the set with the highest set of units by considering the relevance weights assigned by the optimization procedure within  VEBI.
To illustrate this issue, we show the number of identified units per layer by different sample selection for Resnet18 in Fig.~\ref{fig:vebi_unitindices_analysis}.(left) and their intersection coverage with those from complete dataset in Fig.~\ref{fig:vebi_unitindices_analysis}.(right). Intersection coverage refers to the percentage of the units identified from the coreset and the dataset that have the same index (displayed in each cell in Fig.~\ref{fig:vebi_unitindices_analysis}.right). As can be seen the highest  difference comes from the deeper layers. Moreover, since  VEBI covers the entire model, we measure representation  similarity for all layers wherein at least one relevant unit has been identified by both the coreset and complete dataset. The final similarity is reported as the averages of similarities over the considered layers.

\begin{figure*}[t]
  \centering
   \includegraphics[width=\linewidth]{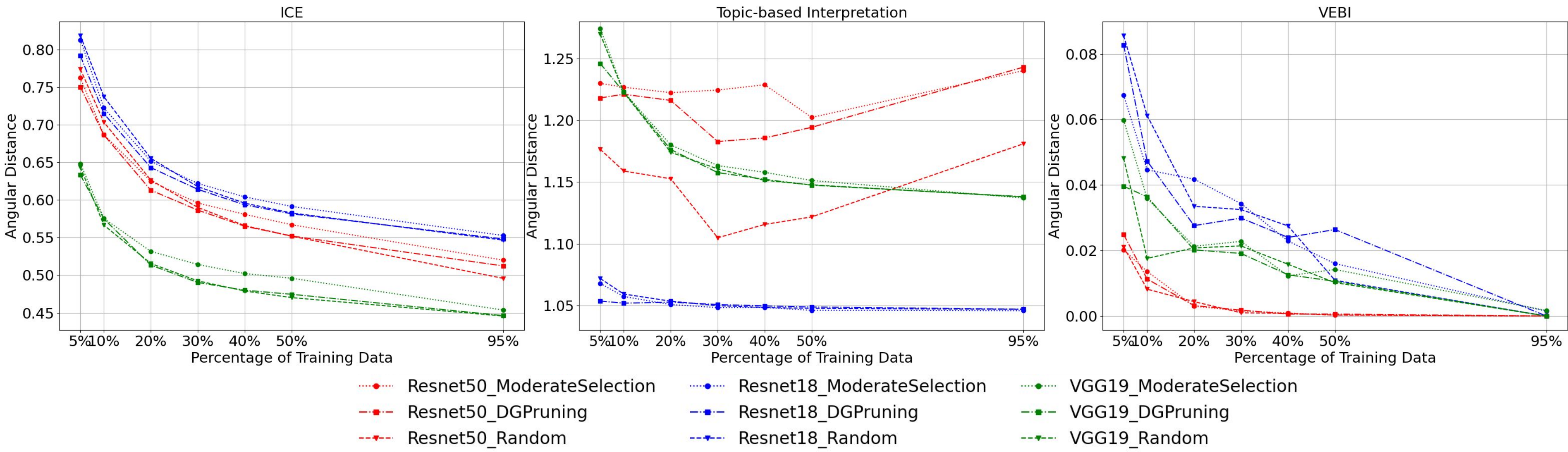}
   \caption{Relevant features similarity between corset and complete dataset for interpretation methods ICE (left), Topic-based interpretation (middle), VEBI (right) over VGG19, Resnet18, and Resnet50.}
   \label{fig:quantitative_repsim}
\end{figure*}

Considering the above settings, we report in Fig.~\ref{fig:quantitative_repsim}, from left to right, the representation similarity for the features identified/learned by  ICE, Topic-based interpretation and VEBI, respectively. In this experiment, we report angular distance. As a result, lower angular distance indicates  higher similarity. In each plot, color represents the considered CNN models and the markers represent the coreset selection methods. 
For reference, and to assess the quality of the considered similarity metric, we report results for a coreset size that  is close to that of the complete dataset, specifically at $95\%$ coreset size.

The following trends can be observed in our results. Firstly, considering a very small data budget, for example $20\%$, the representation similarity between coreset selection and complete dataset is close to the representation similarity when $95\%$ of dataset is considered as coreset. This shows that with a very few data from a large dataset, interpretation methods provide relevant features representing similar characteristics to those from the complete dataset. 
Second, among coreset selection methods and for majority of cases in lower data budget, random selection has comparative results to geometry-based selection methods. This shows random selection is a strong baseline for extracting relevant features in comparison to other selection methods. 
Third, among the considered  interpretation methods, the results from VEBI show the highest similarity, i.e., closest to those from complete dataset. 

In summary, considering the quantitative experiments from model classification section (Sec.~\ref{sec:acc_intrpr}) and coreset-based relevant features similarity in this section, coreset selection methods enable interpretation methods to operate under a limited data budget. Moreover, given the considered interpretation methods, VEBI has better performance via coreset selection methods in comparison to ICE and Topic-based interpretation in terms of providing relevant features via a very limited data that is more similar to those from complete dataset.

\subsection{Robustness of interpretation w.r.t. data budget}
This section investigates the robustness of the considered interpretation methods with respect to the size of the dataset fed to them as input. To do this, we summarize the corset-based relevant features similarity by computing their average and standard deviation across the considered selection ratios $10\%$, $20\%$, $30\%$, $40\%$, and $50\%$ (Fig.~\ref{fig:quantitative_repsim}) which are presented in  Table~\ref{table:interpretaion_stability}.

\begin{table}[]
\caption{Interpretation methods stability in terms of average and standard deviation of coreset-based relevant features similarity, when using different data budget.
}
\centering
\label{table:interpretaion_stability}
\resizebox{\linewidth}{!}{%
\begin{tabular}{|l|c|c|c|}
\hline
 & \multicolumn{3}{c|}{\textbf{VGG19}} \\
\hline
Interpretation Method & VEBI & ICE & Topic-based  \\

Corset Selection& & & Interpretation\\
\hline
Random &%
\textbf{1.71e$-$2$\pm$4.04e$-$3} & 5.04e$-$1$\pm$3.44e$-$2 & 1.17 $\pm$2.72$-$2 \\ 

DGPruning &%
1.97e$-$2$\pm$9.08e$-$3 & 5.06e$-$1$\pm$3.66e$-$2 & 1.17 $\pm$2.72$-$2\\ 

Moderate Selection &%
2.13e$-$2$\pm$8.32e$-$3 & 5.23e$-$1$\pm$2.85e$-$2 & 1.17$\pm$2.58$-$2\\ 
\hline

 & \multicolumn{3}{c|}{\textbf{Resnet18}} \\
\hline
Random &%
3.31e$-$2$\pm$1.62e$-$2 & 6.37e$-$1$\pm$5.54e$-$2 & 1.05$\pm$4.26$-$3 \\ 

DGPruning &%
\textbf{3.10e$-$2$\pm$8.34e$-$3} & 6.29e$-$1$\pm$4.74e$-$2 & 1.05$\pm$1.41$-$3 \\ 
Moderate Selection &%
3.19e$-$2$\pm$1.09e$-$2 & 6.38e$-$1$\pm$4.66e$-$2 & 1.05$\pm$3.85$-$3\\ 
\hline
 & \multicolumn{3}{c|}{\textbf{Resnet50}} \\
\hline
Random &%
\textbf{2.94e$-$3$\pm$3.00e$-$3} & 6.07e$-$1$\pm$5.41e$-$2 & 1.13$\pm$2.12$-2$\\

DGPruning &%
3.49e$-$3$\pm$4.01e$-$3 & 6.00e$-$1$\pm$4.75e$-$2 & 1.20$\pm$1.57$-$2\\

Moderate Selection &%
3.88e$-$3$\pm$4.94e$-$3 & 6.10e$-$1$\pm$4.23e$-$2 & 1.22$\pm$9.57$-3$\\
\hline
\end{tabular}%
}
\end{table}

In Table~\ref{table:interpretaion_stability}, lower angular distance values  indicate  higher similarity and better stability. As evident, Topic-based interpretation has the lowest similarity, i.e., the highest angular distance across all the coreset methods and CNNs. With the exception of Topic-based interpretation on Resnet18 that have the lowest standard deviation, it can be seen that VEBI has the lowest angular distance, i.e., highest similarity, along with the lowest standard deviation across different coreset selection and CNNs. This indicates that VEBI has highest robustness towards different data budgets in comparison to those of ICE and Topic-based interpretation. 

\subsection{Coreset transferability in model interpretation}
\label{sec:coreset_transferability}
\begin{figure}[t]
  \centering
   \includegraphics[width=\linewidth]{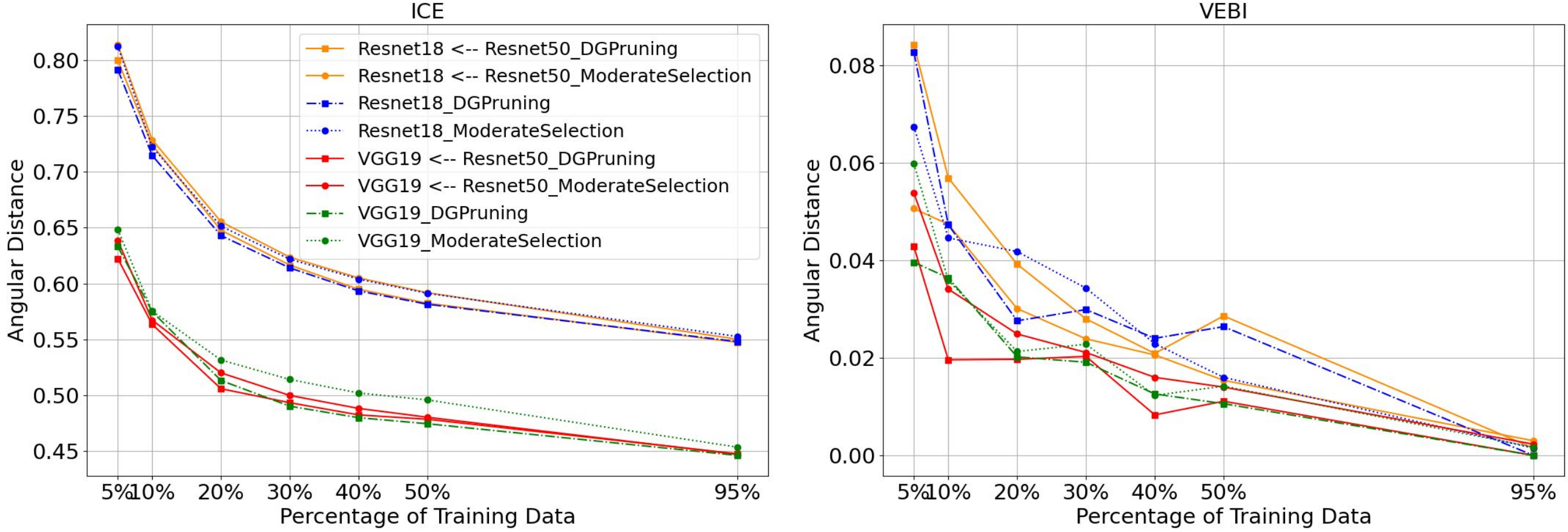}
   \caption{Relevant feature similarity using coreset transferability, i.e., corsets determined by Resnet50, for interpretation methods ICE (top) and VEBI (down) over VGG19 and Resnet18.}
   \label{fig:corset_transferability}
\end{figure}

We investigate the transferability of the coreset obtained from a model for identifying/learning relevant features from other models. To do so, we applied ICE and VEBI on Resnet18 and VGG19 to extract relevant features using coresets determined from activation maps computed by Resnet50. Fig.~\ref{fig:corset_transferability} shows feature similarity with (marked in the legend as  $CNN {<--} Resnet50\_coreset\_method$) and without coreset transferability. As can be seen, the relevant features extracted via the interpretation methods considering the Resnet50-transferred coreset are very close to those obtained from the coreset obtained from those models. This suggests that the coresets are indeed transferable for interpretation task and do not need to be re-extracted for each model, thus reducing the computational cost of their extraction.

\subsection{Computational complexity}
This section briefly discusses the computational cost of each interpretation method when using lower data budget in comparison to the complete dataset.

\textbf{ICE}~\cite{ice}. This method utilizes Non-negative Factorization~\cite{fevotte2011algorithms} to learn relevant features. The computational cost of this method is $O(ndr)$~\cite{fevotte2011algorithms}, where $n$, $d$, and $r$ are the number of samples, feature, and components, respectively, in the factorization procedure, indicating the relevant features. Considering $d$ and $r$ as fixed parameters, the computational complexity is reduced proportionally to the percentage of data $\rho$ (in percentage), i.e., $O\left(\rho nrd\right) < O(ndr)$.

\textbf{VEBI}~\cite{vebi}. This method formulates the identification of internals via a $\mu$-lasso formulation. Solving this problem with the condition of $d{<}n$ and $d^3{<}d^2n$ using the LARS algorithm results in a computational cost of $O(nd^2)$~\cite{efron2004least}.
With fixed $d$ and different data selection ratios $\rho$, the computational cost is further reduced to $O\left(\rho nd^2\right){<}O(nd^2)$.

\textbf{Topic-based interpretation}~\cite{topicmodel}. According to its description and implementation\footnote{Topic-based interpretation method: $https://github.com/chihkuanyeh/concept\_exp$}, we have calculated its computation order to be $O(ndhw(m+2ml))$, where $n$, $d$, $h$, $w$ represent the number of data points, depth, height, and width of the input, respectively. $m$ indicates the number of topics and l shows the length of a trainable parameter in the interpretable module. We have provided the details of this calculations in the Supplementary Material. Similar to those of VEBI, and ICE, given different selection ratios $\rho$, the computational complexity is reduced to $O(\rho ndhw(m+2ml)) < O(ndhw(m+2ml))$. 

In conclusion, observations from Sec.~\ref{sec:coreset_transferability} show that coreset transferability helps in reducing the computational overhead of extracting coresets while preserving the quality of obtained insights. Also, the discussion from this section indicates lower data budget yields in reducing the computation cost of interpretation methods. As a result, the proposed interpretation pipeline based on coreset selection helps in speeding-up of interpretation task.   

\subsection{Interpretation feedback visualization}
This section presents examples of interpretation feedback visualization. 
We compute interpretation maps $\boldsymbol{Z}^i \in \mathbb{R}^{h'\times w'\times d'}$ with elements in $(u,v)$ locations as the result of employing activation map $\boldsymbol{A^i}$ and relevant features $\boldsymbol{R'}$ (Sec.~\ref{sec:framework_setup}). Then, each interpretation map is scored based on the summation of its elements in $(u,v)$ locations.
The $k$ interpretation maps with the highest score are  selected, that is $r {\in} \mathbb{R}^{w' \times h' \times k}$ and combined together to form a heatmap $\boldsymbol{H}^i$ by selecting maximum element in each $(u,v)$ location across the $k$ channels. 
Finally, we resize $\boldsymbol{H}^i {\in} \mathbb{R}^{{w' \times h'}}$ to the size of input sample $\boldsymbol{X}^i$ and superimpose it on the input sample to present regions highlighted by the relevant features $\boldsymbol{R'}$. 

\begin{figure}[t]
  \centering
   \includegraphics[width=\linewidth]{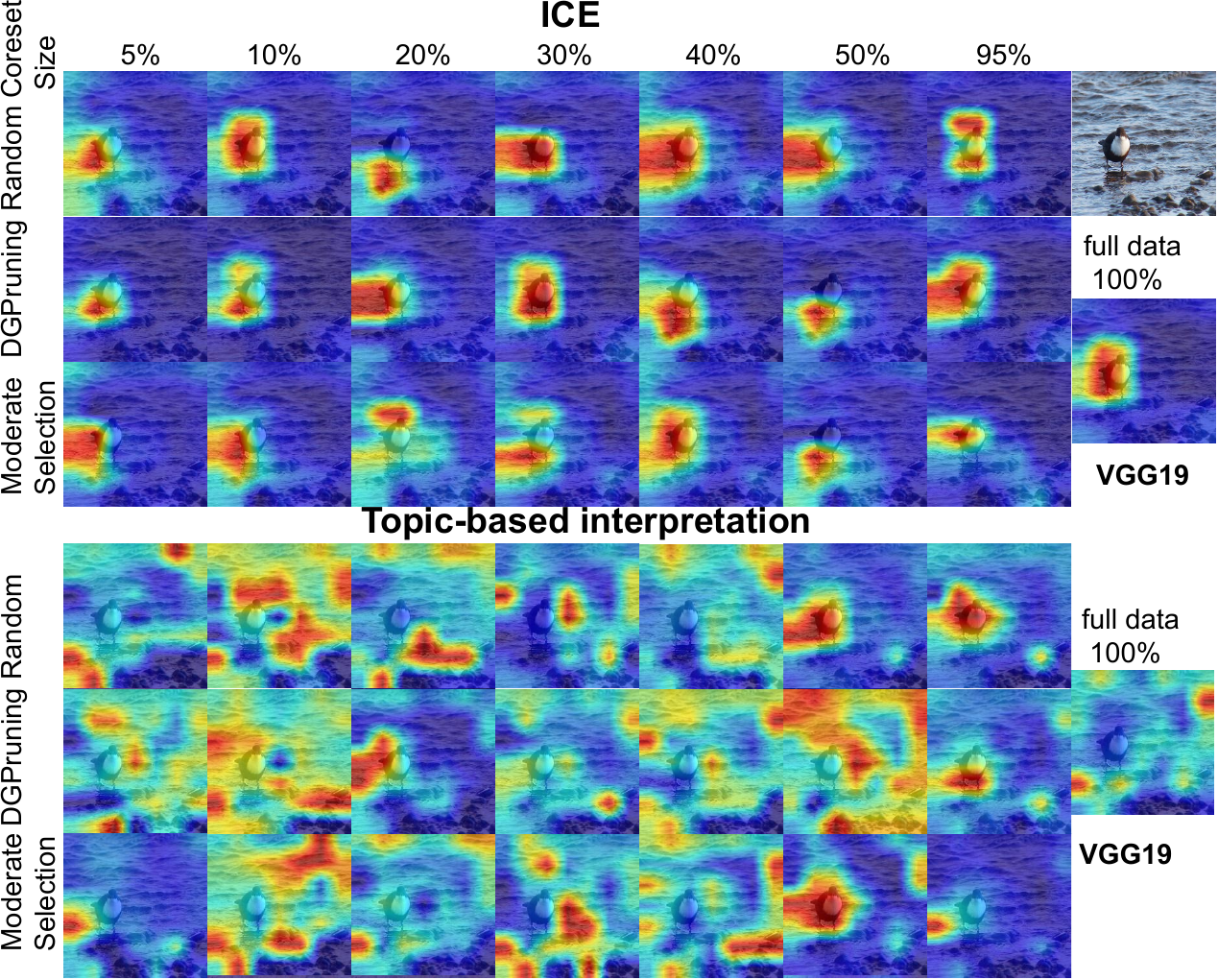}
   \caption{Interpretation feedback visualization from complete dataset and different coreset selection methods for the ICE (\textbf{top}) and Topic-based interpretation (\textbf{bottom}) methods over VGG19.}
   \label{fig:vis_ice_topic_vgg19}
\end{figure}

\begin{figure}[t]
  \centering
   \includegraphics[width=\linewidth]{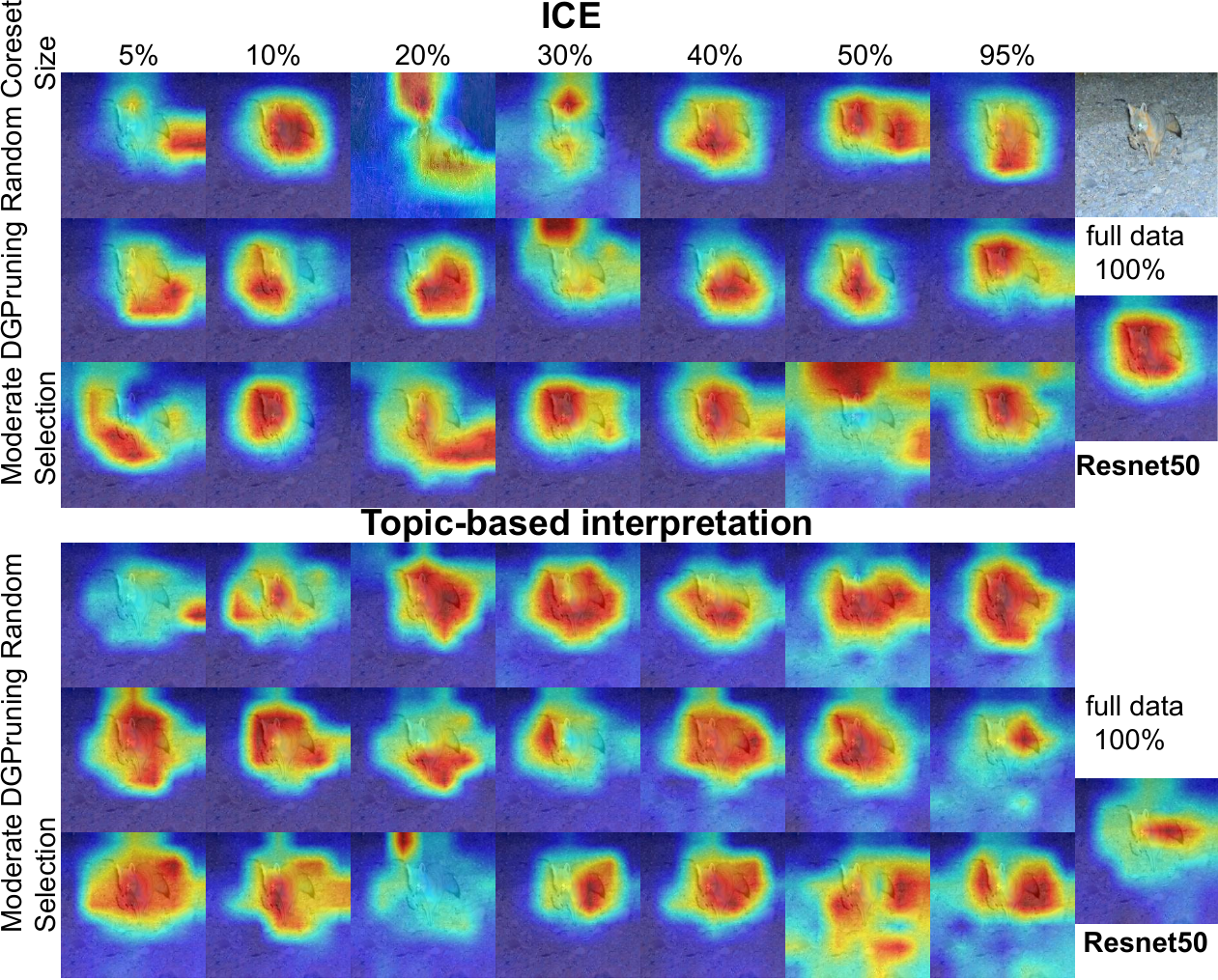}
   \caption{Interpretation feedback visualization from complete dataset and different coreset selection methods for the ICE (\textbf{top}) and Topic-based interpretation (\textbf{bottom}) methods over Resnet50.}
   \label{fig:vis_ice_topic_res50}
\end{figure}

Figures~\ref{fig:vis_ice_topic_vgg19} and \ref{fig:vis_ice_topic_res50} show examples of interpretation feedback visualizations obtained from ICE(top) and Topic-based interpretation(bottom) on VGG19 and Resnet50, respectively. Additional visualizations from other CNNs are available in the Supplementary Material. 
The illustrations on the left to right side depict learned relevant features using each coreset selection method, while the single visualization on the right displays the result obtained from using the full dataset for learned relevant features. It is evident that the visualizations from ICE highlight regions that are qualitatively similar to the highlighted regions in the visualization from the full dataset. In contrast, the visualizations from Topic-based interpretation obtained using different coreset selection methods are  different. This discrepancy is also reflected in the results presented in Fig.~\ref{fig:quantitative_repsim}, where the obtained relevant features via coreset selection methods from ICE is closer to that from the complete dataset across all CNNs, compared to Topic-based interpretation.

\begin{figure}[t]
  \centering
   \includegraphics[width=\linewidth]{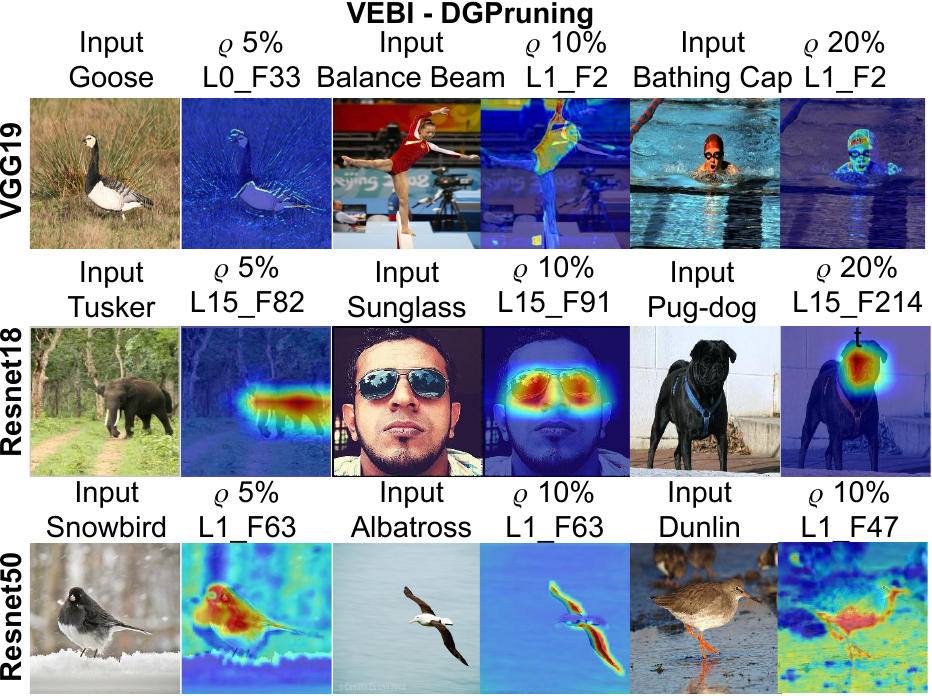}
   \caption{Visualization of identified units by VEBI using DGPruning, different from those identified using complete dataset, over VGG19, Resnet18, and Resnet50.}
   \label{fig:vebi_differentunits_vis}
\end{figure}

Since VEBI identifies relevant units and reports their indices, the same units identified by both the coreset and complete dataset produce identical activation maps. Hence, we focus our qualitative analysis on the different units identified between the coreset and the complete dataset. This will help to inspect whether they highlight regions semantically similar to the visual patterns presented in the class. Fig.~\ref{fig:vebi_differentunits_vis} shows some examples, wherein $\rho$ shows corseset size, \textit{L} and \textit{F} indicate the index of layer and filter, respectively. As can be noted, the differing identified units still focus on visual patterns that are relevant to the classes of interest.

\section{Conclusion}
We present a framework based on coreset selection as a pre-processing step for the model  interpretation task. As part of the proposed method, we introduce a similarity-based evaluation protocol to assess the quality of interpretation obtained by coreset selection. We also investigate the stability w.r.t. the data volumes fed to them. Our results show the effectiveness of the utilized coreset selection in reducing the computational cost of the interpretation methods while preserving the quality of the extracted insights.

{\small
\bibliographystyle{ieee_fullname}
\bibliography{PaperForReview}

\begin{thebibliography}{10}\itemsep=-1pt

\bibitem{linear_probing}
Guillaume Alain and Yoshua Bengio.
\newblock Understanding intermediate layers using linear classifier probes.
\newblock {\em In 5th International Conference on Learning Representations (ICLR), page 68, Toulon, France}, 2016.

\bibitem{unified_protocol}
Hamed Behzadi-Khormouji and Jos{\'e} Oramas.
\newblock A protocol for evaluating model interpretation methods from visual explanations.
\newblock In {\em Proceedings of the IEEE/CVF Winter Conference on Applications of Computer Vision}, pages 1421--1429, 2023.

\bibitem{protopnet}
Chaofan Chen, Oscar Li, Alina Barnett, Jonathan Su, and Cynthia Rudin.
\newblock This looks like that: deep learning for interpretable image recognition.
\newblock {\em Advances in neural information processing systems (NeurIPS)}, 32, 2019.

\bibitem{conceptwhitening}
Zhi Chen, Yijie Bei, and Cynthia Rudin.
\newblock Concept whitening for interpretable image recognition.
\newblock {\em Nature Machine Intelligence}, 2(12):772--782, 2020.

\bibitem{coleman2020selection}
Cody Coleman, Christopher Yeh, Stephen Mussmann, Baharan Mirzasoleiman, Peter Bailis, Percy Liang, Jure Leskovec, and Matei Zaharia.
\newblock Selection via proxy: Efficient data selection for deep learning.
\newblock {\em International Conference on Learning Representations (ICLR)}, 2020.

\bibitem{davari2023reliability}
MohammadReza Davari, Stefan Horoi, Amine Natik, Guillaume Lajoie, Guy Wolf, and Eugene Belilovsky.
\newblock Reliability of cka as a similarity measure in deep learning.
\newblock {\em International Conference on Learning Representations (ICLR)}, 2023.

\bibitem{imagenet}
Jia Deng, Wei Dong, Richard Socher, Li-Jia Li, Kai Li, and Li Fei-Fei.
\newblock Imagenet: A large-scale hierarchical image database.
\newblock In {\em 2009 IEEE Conference on Computer Vision and Pattern Recognition}, pages 248--255, 2009.

\bibitem{network_inversion}
Alexey Dosovitskiy and Thomas Brox.
\newblock Inverting visual representations with convolutional networks.
\newblock In {\em Proceedings of the IEEE conference on computer vision and pattern recognition (CVPR)}, pages 4829--4837, 2016.

\bibitem{efron2004least}
Bradley Efron, Trevor Hastie, Iain Johnstone, and Robert Tibshirani.
\newblock Least angle regression.
\newblock 2004.

\bibitem{fevotte2011algorithms}
C{\'e}dric F{\'e}votte and J{\'e}r{\^o}me Idier.
\newblock Algorithms for nonnegative matrix factorization with the $\beta$-divergence.
\newblock {\em Neural computation}, 23(9):2421--2456, 2011.

\bibitem{net2vec}
Ruth Fong and Andrea Vedaldi.
\newblock Net2vec: Quantifying and explaining how concepts are encoded by filters in deep neural networks.
\newblock In {\em Proceedings of the IEEE conference on computer vision and pattern recognition}, pages 8730--8738, 2018.

\bibitem{nmf_survey}
Jiangzhang Gan, Tong Liu, Li Li, and Jilian Zhang.
\newblock Non-negative matrix factorization: a survey.
\newblock {\em The Computer Journal}, 64(7):1080--1092, 2021.

\bibitem{intervention}
Itai Gat, Guy Lorberbom, Idan Schwartz, and Tamir Hazan.
\newblock Latent space explanation by intervention.
\newblock In {\em Proceedings of the AAAI Conference on Artificial Intelligence}, volume~36, pages 679--687, 2022.

\bibitem{ace}
Amirata Ghorbani, James Wexler, James Zou, and Been Kim.
\newblock Towards automatic concept-based explanations.
\newblock {\em Advances in Neural Information Processing Systems (NeurIPS)}, 32, 2019.

\bibitem{deepcore}
Chengcheng Guo, Bo Zhao, and Yanbing Bai.
\newblock Deepcore: A comprehensive library for coreset selection in deep learning.
\newblock In {\em International Conference on Database and Expert Systems Applications}, pages 181--195. Springer, 2022.

\bibitem{resnet50}
Kaiming He, Xiangyu Zhang, Shaoqing Ren, and Jian Sun.
\newblock Deep residual learning for image recognition.
\newblock In {\em Proceedings of the IEEE conference on computer vision and pattern recognition}, pages 770--778, 2016.

\bibitem{pace}
Vidhya Kamakshi, Uday Gupta, and Narayanan~C Krishnan.
\newblock Pace: Posthoc architecture-agnostic concept extractor for explaining cnns.
\newblock In {\em 2021 International Joint Conference on Neural Networks (IJCNN)}, pages 1--8. IEEE, 2021.

\bibitem{tcav}
Been Kim, Martin Wattenberg, Justin Gilmer, Carrie Cai, James Wexler, Fernanda Viegas, et~al.
\newblock Interpretability beyond feature attribution: Quantitative testing with concept activation vectors (tcav).
\newblock In {\em International conference on machine learning (ICML)}, pages 2668--2677. PMLR, 2018.

\bibitem{revers_probing}
Iro Laina, Yuki~M. Asano, and Andrea Vedaldi.
\newblock Measuring the interpretability of unsupervised representations via quantized reverse probing.
\newblock In {\em International Conference on Learning Representations (ICLR).}, 2022.

\bibitem{d2pruning}
Adyasha Maharana, Prateek Yadav, and Mohit Bansal.
\newblock D2 pruning: Message passing for balancing diversity and difficulty in data pruning.
\newblock {\em In International Conference on Learning Representations (ICLR)}, 2024.

\bibitem{acts_mining2023}
Toon Meynen, Hamed Behzadi-Khormouji, and Jos{\'e} Oramas.
\newblock Interpreting convolutional neural networks by explaining their predictions.
\newblock In {\em 2023 IEEE International Conference on Image Processing (ICIP)}, pages 1685--1689. IEEE, 2023.

\bibitem{prtotree}
Meike Nauta, Ron van Bree, and Christin Seifert.
\newblock Neural prototype trees for interpretable fine-grained image recognition.
\newblock In {\em Proceedings of the IEEE/CVF Conference on Computer Vision and Pattern Recognition (CVPR)}, pages 14933--14943, 2021.

\bibitem{nguyen2020wide}
Thao Nguyen, Maithra Raghu, and Simon Kornblith.
\newblock Do wide and deep networks learn the same things? uncovering how neural network representations vary with width and depth.
\newblock {\em arXiv preprint arXiv:2010.15327}, 2020.

\bibitem{vebi}
Jos{\'{e}} Oramas, Kaili Wang, and Tinne Tuytelaars.
\newblock Visual explanation by interpretation: Improving visual feedback capabilities of deep neural networks.
\newblock {\em In International Conference on Learning Representations (ICLR)}, 2019.

\bibitem{pytorch}
Adam Paszke, Sam Gross, Francisco Massa, Adam Lerer, James Bradbury, Gregory Chanan, Trevor Killeen, Zeming Lin, Natalia Gimelshein, Luca Antiga, Alban Desmaison, Andreas Kopf, Edward Yang, Zachary DeVito, Martin Raison, Alykhan Tejani, Sasank Chilamkurthy, Benoit Steiner, Lu Fang, Junjie Bai, and Soumith Chintala.
\newblock Pytorch: An imperative style, high-performance deep learning library.
\newblock In {\em Advances in Neural Information Processing Systems 32}, pages 8024--8035. Curran Associates, Inc., 2019.

\bibitem{paul2021deep}
Mansheej Paul, Surya Ganguli, and Gintare~Karolina Dziugaite.
\newblock Deep learning on a data diet: Finding important examples early in training.
\newblock {\em Advances in Neural Information Processing Systems}, 34:20596--20607, 2021.

\bibitem{selectivity_index}
Ivet Rafegas, Maria Vanrell, Lu{\'\i}s~A Alexandre, and Guillem Arias.
\newblock Understanding trained cnns by indexing neuron selectivity.
\newblock {\em Pattern Recognition Letters}, 136:318--325, 2020.

\bibitem{raghu2021vision}
Maithra Raghu, Thomas Unterthiner, Simon Kornblith, Chiyuan Zhang, and Alexey Dosovitskiy.
\newblock Do vision transformers see like convolutional neural networks?
\newblock {\em Advances in Neural Information Processing Systems}, 34:12116--12128, 2021.

\bibitem{protopshare}
Dawid Rymarczyk, {\L}ukasz Struski, Jacek Tabor, and Bartosz Zieli{\'n}ski.
\newblock Protopshare: Prototypical parts sharing for similarity discovery in interpretable image classification.
\newblock In {\em Proceedings of the 27th ACM SIGKDD Conference on Knowledge Discovery \& Data Mining (KDD)}, pages 1420--1430, 2021.

\bibitem{sener2018active}
Ozan Sener and Silvio Savarese.
\newblock Active learning for convolutional neural networks: A core-set approach.
\newblock {\em International Conference on Learning Representations (ICLR)}, 2018.

\bibitem{vgg19}
Karen Simonyan and Andrew Zisserman.
\newblock Very deep convolutional networks for large-scale image recognition.
\newblock {\em In International Conference on Learning Representations (ICLR)}, 2015.

\bibitem{sorscher2022beyond}
Ben Sorscher, Robert Geirhos, Shashank Shekhar, Surya Ganguli, and Ari Morcos.
\newblock Beyond neural scaling laws: beating power law scaling via data pruning.
\newblock {\em Advances in Neural Information Processing Systems}, 35:19523--19536, 2022.

\bibitem{toneva2018empirical}
Mariya Toneva, Alessandro Sordoni, Remi Tachet~des Combes, Adam Trischler, Yoshua Bengio, and Geoffrey~J Gordon.
\newblock An empirical study of example forgetting during deep neural network learning.
\newblock {\em International Conference on Learning Representations (ICLR)}, 2018.

\bibitem{wang2018towards}
Liwei Wang, Lunjia Hu, Jiayuan Gu, Zhiqiang Hu, Yue Wu, Kun He, and John Hopcroft.
\newblock Towards understanding learning representations: To what extent do different neural networks learn the same representation.
\newblock {\em Advances in neural information processing systems}, 31, 2018.

\bibitem{subnetworks}
Yulong Wang, Hang Su, Bo Zhang, and Xiaolin Hu.
\newblock Interpret neural networks by extracting critical subnetworks.
\newblock {\em IEEE Transactions on Image Processing}, 29:6707--6720, 2020.

\bibitem{williams2021generalized}
Alex~H Williams, Erin Kunz, Simon Kornblith, and Scott Linderman.
\newblock Generalized shape metrics on neural representations.
\newblock {\em Advances in Neural Information Processing Systems}, 34:4738--4750, 2021.

\bibitem{concept_attribution}
Weibin Wu, Yuxin Su, Xixian Chen, Shenglin Zhao, Irwin King, Michael~R Lyu, and Yu-Wing Tai.
\newblock Towards global explanations of convolutional neural networks with concept attribution.
\newblock In {\em Proceedings of the IEEE/CVF Conference on Computer Vision and Pattern Recognition (CVPR)}, pages 8652--8661, 2020.

\bibitem{moderate_selection}
Xiaobo Xia, Jiale Liu, Jun Yu, Xu Shen, Bo Han, and Tongliang Liu.
\newblock Moderate coreset: A universal method of data selection for real-world data-efficient deep learning.
\newblock In {\em The Eleventh International Conference on Learning Representations}, 2022.

\bibitem{topicmodel}
Chih{-}Kuan Yeh, Been Kim, Sercan~{\"{O}}mer Arik, Chun{-}Liang Li, Pradeep Ravikumar, and Tomas Pfister.
\newblock On concept-based explanations in deep neural networks.
\newblock {\em In International Conference on Learning Representations (ICLR)}, 2020.

\bibitem{ice}
Ruihan Zhang, Prashan Madumal, Tim Miller, Krista~A Ehinger, and Benjamin~IP Rubinstein.
\newblock Invertible concept-based explanations for cnn models with non-negative concept activation vectors.
\newblock In {\em Proceedings of the AAAI Conference on Artificial Intelligence}, volume~35, pages 11682--11690, 2021.

\bibitem{networkdissection}
Bolei Zhou, David Bau, Aude Oliva, and Antonio Torralba.
\newblock Interpreting deep visual representations via network dissection.
\newblock {\em IEEE transactions on pattern analysis and machine intelligence}, 41(9):2131--2145, 2018.

\end{thebibliography}
}

\end{document}